# TRAINING A FEED-FORWARD NEURAL NETWORK WITH ARTIFICIAL BEE COLONY BASED BACK-PROPAGATION METHOD

Sudarshan Nandy[1] , Partha Pratim Sarkar[2] and Achintya Das[3]

[1&2] DETS, Kalyani University,
Kalyani, Nadia, West Bengal, India.
[1]sudarshannandy@gmail.com and [2]parthabe91@yahoo.co.in
[3]Kalyani Goverment Engineering College,
Kalyani, Nadia, West Bengal, India.
achintya.das123@gmail.com

## *ABSTRACT*

*Back-propagation algorithm is one of the most widely used and popular techniques to optimize the feed forward neural network training. Nature inspired meta-heuristic algorithms also provide derivative-free solution to optimize complex problem. Artificial bee colony algorithm is a nature inspired meta-heuristic algorithm, mimicking the foraging or food source searching behaviour of bees in a bee colony and this algorithm is implemented in several applications for an improved optimized outcome. The proposed method in this paper includes an improved artificial bee colony algorithm based back-propagation neural network training method for fast and improved convergence rate of the hybrid neural network learning method. The result is analysed with the genetic algorithm based back-propagation method, and it is another hybridized procedure of its kind. Analysis is performed over standard data sets, reflecting the light of efficiency of proposed method in terms of convergence speed and rate.*

## *KEYWORDS*

*Neural Network, Back-propagation Algorithm, Meta-heuristic algorithm.*

## 1. INTRODUCTION

The use of nature inspired meta-heuristic algorithm to solve many complex problems is increasingly gaining its popularity in scientific community. In recent year various conventional optimization algorithms, inspired by nature, are developed to meet the challenge of many complex applications. The main job of those algorithm is to provide a derivative free local search. Nature inspired algorithm is initiated with a guess on its parameter value, and the value of the parameter is modified on the basis of numerical analysis of algorithmic performance. Those meta-heuristic algorithms are initiated with randomly generated population which consists of a solution to the given problem, and parameters are modified according to the fitness of the solution. Algorithms that are categorized into this population based methods are genetic algorithm[1], ant colony optimization[2], differential evaluation algorithm[3], particle swarm optimization[4], artificial bee colony algorithm[5] etc. The genetic algorithm is the most widely used evolutionary algorithm, and it is implemented and tested over several complex optimization problem. This algorithm is suffered from its slow convergence speed due to large searching space[6]. The selection of parent in this algorithm totally depends on the fitness value, and hence there is no chance for other to get selected as a parent. In differential evolution algorithm, the chance of being selected as a parent is equal for all the solutions. The main difference between genetic and differential algorithm is the parent selection procedure. Instead of many advantages, the differential algorithm and genetic algorithm suffer from the Hamming cliff problem[7,8]. Ant and particle swarm optimization are also used for many complex optimizations, and both of these algorithms are studied on various complex applications[9,10]. Artificial bee colony algorithm is inspired by the bee behavior in the bee colony algorithm. The individual efforts of collecting nectar and performing various complex tasks like understanding the signal pattern, following other bees to find a source of food and remembering old food source influence the algorithm. Various modified versions of this algorithm can be found which making it most perfectly tuned for complex optimization problem. Among many implementations of this algorithm, the bee colony optimization based routing wavelength assignment problem[12], the dynamic allocation of Internet service[11] ,the bee algorithm based energy efficient routing for ad-hoc mobile network[13] etc. are the few complex and successful applications.’

The back-propagation algorithm is one of the most famous algorithm to train a feed forward network. Instead of its success rate the quest for development is observed through the various standard modifications in-order to meet the challenges of complex applications. The development phase of this algorithm is generally categorized into two specific branches. In the first branch the considerable research on ad-hoc modification of parameter is observed. This is generally known as a heuristic approach towards the back-propagation training optimization[14,15,16]. This kind of algorithm is designed with variable learning rate where variations in learning rate depends on the performance index. Modification of several other parameters that are responsible to converge the algorithm, is also hinge on performance index of the algorithm. Example of this type of algorithms are variable learning back-propagation, momentum based back-propagation. Numerical optimization techniques are incorporated into back-propagation algorithm which is categorized for this type of development into another phase. In this phase some of the techniques produce an accelerated result on convergence for learning, and those algorithms are not derivative-free for optimization process. The main advantages of this type of algorithms are easy to implement, and parameter adjustment is not required for this algorithm. Conjugate back-propagation[17] and Levenberg-Marquardt back-propagation [18,19,20] methods are widely used algorithm in this category. Apart from wide range of successful implementation result these algorithms require more memory storage, computation, and there is always a risk of grapple with local minimum, as they are not derivative free.

Another significant effort is found in the development of hybrid method of optimizing back-propagation training. This hybridization is caused by incorporating two nature inspired methods. Genetic algorithm based back-propagation [6] is one kind of hybridized method to optimize the neural network training. This method is slow in convergence because of its large search space. In this population based approach, the algorithm use fitness value of solution as a performance index. The other methods effectively applied to back-propagation for training optimization are artificial bee colony algorithm and particle swarm optimizations[21,22,23]. In some cases it is observed that the hybridization method implemented in the back-propagation, training optimization updates the weight of each node in a neural network by implementing the Leveberg-Marquardt or any other standard optimization algorithm in training phase, and hence artificial bee colony method is indirectly responsible to modify the weight and bias of each neuron in optimization phase of back-propagation training. This may cause lot computation power, memory storage. As the non-linear least square method is involved in the optimization process for which this hybridization is not derivative-free, and so, there is a risk of tackling local minimum.

In this proposed method modified artificial bee colony algorithm is implemented inside the optimization phase of back-propagation method, and the same is also used to update weight and bias of each layer in a multilayer feed forward neural network. The foraging behavior of bee and the maintenance of unemployed and employed bees ratio based on the richness of nectar in the honey bee colony are the most important factors to design this meta-heuristic algorithm. The collective intelligence and information sharing nature of honey bees can be easily compared with the group of agent working together to complete a job which actually influence the modification and implementation of the artificial bee colony algorithm in back-propagation neural network training. The result is analyzed with five standard data sets. It is observed that the acceleration in convergence rate is achieved with less number of population, and stability is confirmed within very few iteration.

## 2. ARTIFICIAL BEE COLONY ALGORITHM

The honey bee is categorized into 178 number of species, and there are ten genus-group names[24]. The social structure, quality of nectar and fertilization of crops are the few most important properties for making these species most famous. Due to various geographical locations and differences in weather pattern, the honey bees are found with different color, shape and nature, but irrespective of those differences there are some basic jobs that they perform on daily basis. The foraging or finding new food source and information sharing about new source are the most common jobs. The selection of the food source depends on many parameters. The food source is selected on the basis of the quality of nectar, distance of food source from the colony, quantity of nectar. Apart from food sources the honey bees are categorized according to their assignment. There are two main types of bee found in the honey bee colony such as : 1) Employed bee and 2) unemployed bee. The unemployed bee is one type of bee which doesn't know the food source location. This type of bee generally search the food source randomly or attend the food source from the knowledge of waggle dance. The scout bee is one kind of unemployed bee which starts the searching of a food source without its knowledge. Appearance of this kind of bee generally varies from 5-30% of the total population. The other type of unemployed bee is the onlooker bee which starts to find the food source by the knowledge gained from the waggle dance. The employed bees are the successful onlooker bee with the knowledge of food source. The employed bees generally share the food source information with the other bee, and guide others about richness and the direction of food source. In reality the guidance of employed bees is reflected through their waggle dance in the specific dance area, and this dancing area is the main information sharing center for all employed honey bees. The onlookers bee is the most informative bee, as all the information about food source is available in dancing area. On the basis of available information in the dancing area the onlookers bee selects the best one. Sharing information also depends on the quantity of food, and hence the recruitment of honey bee depends on the quantity and richness of food source. When nectar amount is decreased, the employed bee becomes an unemployed bee, and abandon the food source. The algorithmic process of artificial bee colony simulates the real life scenario of searching a food source, and maintain various types of bees involved in the searching and collecting the nectar. The population of the algorithm represents the honey bee involved in the process of collecting nectar, and it can be expressed as follows:

$$f(x)=\{1,2,\ldots\ldots,N\}$$

Where 'N' is the number of population and f(x) is the objective function. The exploitation of this meta-heuristic algorithm is performed by moving that population towards the food source, and the exploration is achieved by opening the chance to find a new food source. The food source here is treated as the solution for the problem, for which the algorithm is applied. In real world of honey bee, the exploitation is performed by the onlookers, and employed bee, while the exploitation is the responsibility of scout bee. The whole algorithmic process is described as follows:

1. If the employed one remembers the old food source, then it moves towards that food source and collects the nectar. The employed one is moved to the neighbor food source by means of visual information, and evaluate the richness of its nectar. In the algorithm, this step is simulated by means of moving employed bee to the food source, and then evaluate the nectar. Now the movement of the bee can be expressed as:

$$(\beta_{ij}+1)=\beta_{ij}+\varphi(\beta_{ij}-\beta_{ik}) \qquad (1)$$

2. The onlookers bee in the honey bee colony select most probable good food source, if there is more than one waggle dance performed in the dance area. The probability of selection of good food source by the onlookers bee after waggle dance always depends on the food source quality, quantity and distance between food source and bee colony. The probability of selection of food source by the onlookers bee is expressed as:

$$p_i=\frac{S_{ij}}{\sum_{i=1}^{N} S_{ij}} \qquad (2)$$

$S_{ij}$ is the strength of waggle dance of $i^{th}$ bee and $p_{ij}$ is the probability to determine a food source by the onlookers bee.

3. The scout bee is always involved in the exploration of new food source. Thus the need of exploration goes through the discovery of new food source. This is observed that in the artificial bee colony algorithm, the exploration and exploitation are not dependent on each other, and the new exploration always remembers if that source is a energy efficient food source.

## 3. PROPOSED METHODOLOGY

In the proposed method the conventional method of artificial bee colony algorithm is modified by maintaining the basic concept of the algorithm. The modification is implemented in the optimization phase of the back-propagation algorithm. The conventional idea of the exploration and exploitation of food source is same for the proposed method. The food source represents the good optimized solution to the neural network. The quality of the food source represents the performance index of all the solutions. The population generated at the beginning of the process is the number of bee involved as a foragers. Now the responsibility of the scout bee is to find a good food resource, and in the present algorithm only one solution is considered with the performance index having low is considered as best food source. The evaluation of the food sources that are discovered by the randomly generated populations is done through the back-propagation procedure. Now after completion of the first cycle the performance of each solution is kept in an index. In order to generate the population of back-propagation, the weight must be created, and each value of the weight matrix for every layer is randomly generated within the range of [1,0]. Each set of weight and bias is used to create one element in a whole population. The size of the population in the proposed method is fixed to N.

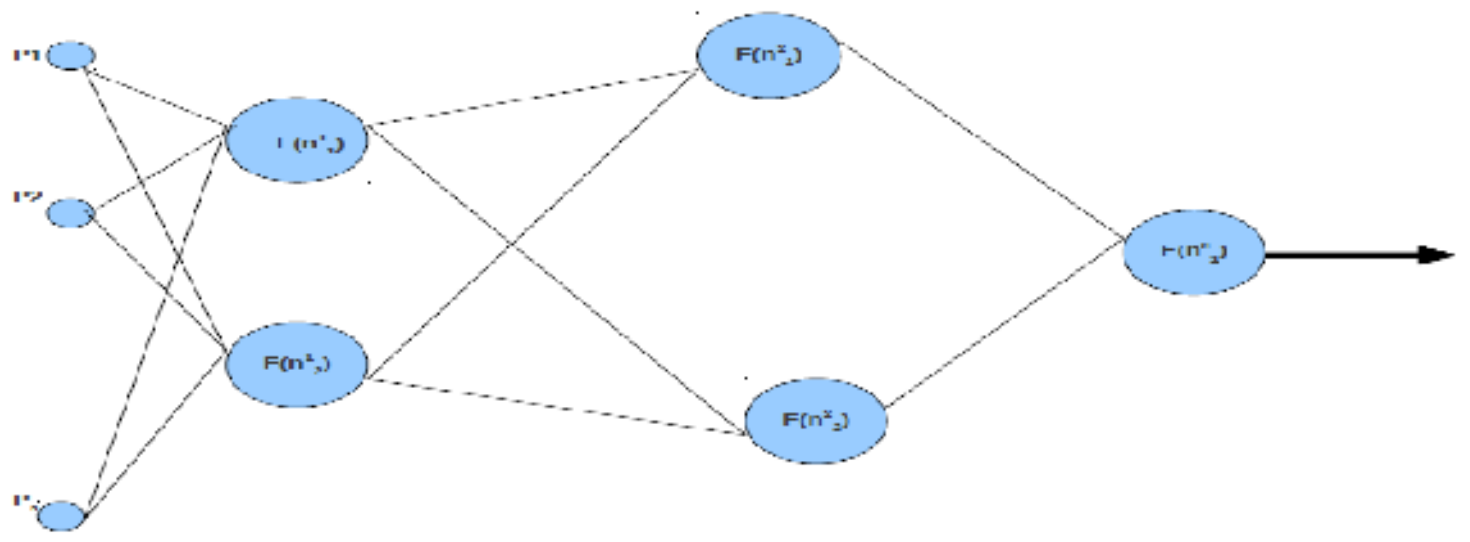

Figure 1: Feed Forward Neural Network Architecture

The list of weight and bias is expressed as:

$$WV_{Li}\,\varepsilon\{rand(0,1)\} \quad \text{and} \quad BV_{Li}\,\varepsilon\{rand(0,1)\}$$

Where, $WV_{Li}$ and $BV_{Li}$ are the elements of a weight and bias matrix. The proposed method uses the feed forward neural network to simulate the back-propagation algorithm (figure 1). The output of the final layer is calculated using following equation:

$$N^{(l+1)}(m) = \sum_{n=1}^{j}\left(W^{(l+1)}(m,n)a_n + B^{(l+1)}\right) \tag{3}$$

Where the N = net output; l= {1,2,3...,j } are the layers; W = weight of the neuron ; B = bias of the neuron. The output is calculated as :

$$a_j(m) = F_j(N_j(m)) \tag{4}$$

where the $a_j$ and $F_j$ are the final output and the transfer function respectively for the $j^{th}$ layer. In the back-propagation process, the sum of squared error is the performance index. The sum of squared error is needed to be calculated for every input data set to understand the performance of learning. In the proposed method the performance index or the fitness value of each population is the average sum of squared error. This is expressed as follows:

$$P_\mu(x) = \frac{\sum_{j=1}^{N} P_F(x)}{N} \tag{5}$$

Where the average sum of squared error or final performance index is $P_\mu(x)$ and the performance Index is $P_F(x)$ and population size is 'N'.

$$p_f(x) = \sum_{i=1}^{l}(t_i - p_i)^T.(t_i - p_i)$$
$$p_f(x) = \sum_{i=1}^{l} e^T.e \tag{6}$$

Where $p_f(x)$ = sum of squared error, $t_i$ = $i^{th}$ target, $p_i$= $i^{th}$ input, e = error
Now , The performance index $P_F(x)$ is calculated as follows:

$$P_F(x) = \sum_{k=1}^{q} p_f(x)^T.p_f(x) \tag{7}$$

So, the first-order derivative of the performance index is as follows:

$$\nabla P_F(x) = \left[ \frac{\delta(p_f(x_1))}{\delta x_1}, \frac{\delta(p_f(x_2))}{\delta x_2}, \ldots \frac{\delta(p_f(x_N))}{\delta x_N} \right] \quad (8)$$

$$\nabla P_F(x) = 2\sum_{(i=1)}^{(N)} p_f(x). \frac{\delta p_f(x)}{\delta x} \quad (9)$$

The weight and bias for the back-propagation is calculated as :
for weight:

$$W_{(l,k)}^{(l+1)} = W_{(l,k)}^{l} - \alpha.S^{l}\left(a^{(l-1)}\right)^{T} \quad (10)$$

and for bias:

$$B_{(l,k)}^{(l+1)} = B_{(l,k)}^{l} - \alpha.S^{l} \quad (11)$$

Where, $\alpha$ = learning rate, $S^l$ = $l^{th}$ layer sensitivity
The sensitivity of one layer is calculated from the sensitivity of the previous layer in a recursive order. The sensitivity of the last layer is calculated as:

$$S^{l} = 2.f_l(N_l).e_i \quad (12)$$

The sensitivity of the other layer is calculated as:

$$S^{(l+1)} = f^{l}(N^{l}).\left(W^{(l+1)}\right)^{T}.S^{(l+1)} \quad (13)$$

The value of $P_\mu(x)$ is the evaluated quality of an element in a population. Now in the proposed method the population must have one scout bee to discover the new food source, and it is considered in this proposed method to maintain the exploration of new food source. Then the food source found by the scout bee is expressed as:

$$F_i = min\{P_\mu(x_1), P_\mu(x_2), P_\mu(x_3), \ldots, P_\mu(x_N)\} \quad (14)$$

Where $F_i$ is the best food source found by the scout bee at the initial stage of the proposed method. At this condition the other population is also observed by the onlookers bee, and it always follows the best possible solution. This onlooker bee is then converted to the employed bee. The probability of selection of food source by the onlooker is expressed as:

$$PF_i = \frac{F_i.d_{ij}}{\left(\sum_{j=1}^{m} F_i\right).d_{ij}} \quad (15)$$

where the $PF_i$ = probability of selecting the solution,

$d_{ij} = \sqrt{(F_i - F_j)^2}$ is the distance between the $F_i$ and $F_j$ ,

$F_j$ = other food source,

j =(1,2,3....,m) is the number of employed to find a food source.

Now according to the proposed method the movement of the employed or onlooker bees to the food source is calculated as:

$$\Delta Fj = \left(F_i - F_j\right) + e^{\left(\cos\left(\frac{Fi}{Fj}\right).\pi\right)} - \log\left(F_i.F_j\right) \quad (16)$$

In the proposed method the movement ($\Delta F_j$) of the employed bee to the food source represents the development of individual solution to the optimum solution. The selection procedure of finding the optimum solution among many others is calculated by the Eqn.(15). In back-propagation neural network training, the modification of weight and bias are the parts of optimization phase calculated as:

$$W_{(l,k)}^{(l+1)} = W_{(l,k)}^{(l)} - \Delta F_j.\alpha \quad (17)$$

and

$$B_{(l,k)}^{(l+1)} = B_{(l,k)}^{(l)} - \Delta F_j.\alpha \quad (18)$$

$P_\mu(x)$ is evaluated when all the data sets are presented through the input of a feed forward neural network for a population. The sum of squared error calculated on each iteration is recalculated if sum of squared error is high or near to infinity. In such cases the modified weight and bias is re-modified accordingly. High sum of squared error represents the low quality food source and hence the employed bee abandon the food source. The re-modification of its weight and bias value to the previous weight and bias value represents that the employed bee became unemployed. According to the performances of the populations the nectar richness is decided through the correct classification rate of neural network learning. The total number of employed bee is decided on the basis of richness factor, and for the proposed methodology it is the correct classification rate. The algorithm terminates its further optimization if the correct classification rate overcomes a pre-specified threshold value. Following is the algorithm for proposed methodology:

*Artificial Bee Colony based Back-propagation*

1. Begin:
2. Random generation of weight and bias of up to a population size N
3. Evaluate the population through back-propagation and store the performance index for each.
4. Find at least one food source $F_i$ at the initial stage using Eqn.(14).
5. initialize i = (0,1,2,....) and maximum cycle number(MCN) = J
6. While (i < MCN):
7. If more than one food source information is available for onlookers bee then calculate the probability of all food source using Eqn.(15).
8. Then move all the onlookers ($F_j$) to the selected food source using Eqn.(16) and calculate the modified weight and bias using Eqn. (17) and Eqn.(18), for every onlookers.
9. Calculate the sum of squared error $P_F(X)$ using Eqn.(7) and stored the result. If the sum of squared error is very high or near to infinity then the weight and bias modification revert back to the previous value and again evaluates the performance.
10. Calculate the average performance index $P_\mu(x)$ and average Correct classification rate of all the population.
11. Now, if the new $P_\mu(x)$ is better than that of the old $P_\mu(x)$ then decreases the number of employed bee and if the old $P_\mu(x)$ is better than that of the new one then remove the present modification and remember the old one and also increases the number scout bee by one.
12. if (correct classification rate < threshold ) then
13. go to step 7 and i += 1
14. else if (correct classification rate > threshold ) then
15. i = MCN (to terminate loop) and go to Step 17
16. else: pass
17. End.

The population for the proposed artificial bee colony based back-propagation optimization needs to be adjusted to get a good convergence rate for every problem. In most of the experiment, 10 numbers of population are used. At the time of initialization, one best solution among other evaluated solutions, is considered as a best solution, and that is found by the scout bee. It is only considered to initialize the proposed algorithmic procedure, and then the number of food source available to be selected by the onlooker bee depends on the correct classification rate. The proposed method of artificial bee colony algorithm based feed-forward back-propagation neural network training converges within a few number of iterations, and it also produces an accelerated convergence rate. The genetic algorithm based back-propagation and proposed one is implemented and analysed using python programming language.

# 4. EVALUATION

The performance of proposed artificial bee colony based back-propagation method is analysed on the basis of correct classification rate and mean of squared error. The result of the analysis is compared with the genetic algorithm based back-propagation method. Four standard data sets are considered here for analysis. Standard data sets available from UCI machine learning library is used to test the performance of proposed method and the genetic algorithm based back-propagation algorithm. The experiment is performed on a computer system with 2.53GHz core2Duo Intel pentium4 processor and it also consists of 3gb of RAM space.

## 4.1. Experimental Data Set

The experiment on the proposed method consists of two phases. In the first part a non-linear standard data sets are used with proposed method for experimental analysis, and in the second part same data sets are used to test the performance of genetic algorithm based back-propagation method. Following data set are used:

Iris data: In this data set there are 150 instances and each instance consists of 4 attributes. There are 3 type of iris flower classes. Those classes are decided by means of their sepal and petal length and width[25].

Wine Data: The wine data set are consists of 178 number of instance and there are 13 attributes in each instance. The win data sets are all about to understand the different quality of wine by means of chemical analysis on three classes of wine[26].

Glass Data: The glass data set consists of seven different type of glass data. This data set consists of 214 number of instance and each instance is made from 10 numbers of glass attributes[27].

Soybean Data: This data set consists of total 47 number of instance and each instance is configured with the 35 number of attributes. The soybean data sets consists of four classes[28].

## 4.2. Experimental Result Analysis

The proposed method is compared with the genetic algorithm based back-propagation (GABPNN) method. In-order to analyse the algorithm, maximum cycle number is set to 100. The neural network used for the back-propagation consists of three layer, and each layer contains one neuron. The learning rate used for the experiment is 0.5. In the proposed artificial bee colony based back-propagation10 numbers of randomly generated population is used for all data sets. Sum of squared error and correct classification rate is used as a convergence criteria. The correct classification rate is divided into maximum, minimum and stable categories. The stable categories represent the classification stability within a fixed iteration.

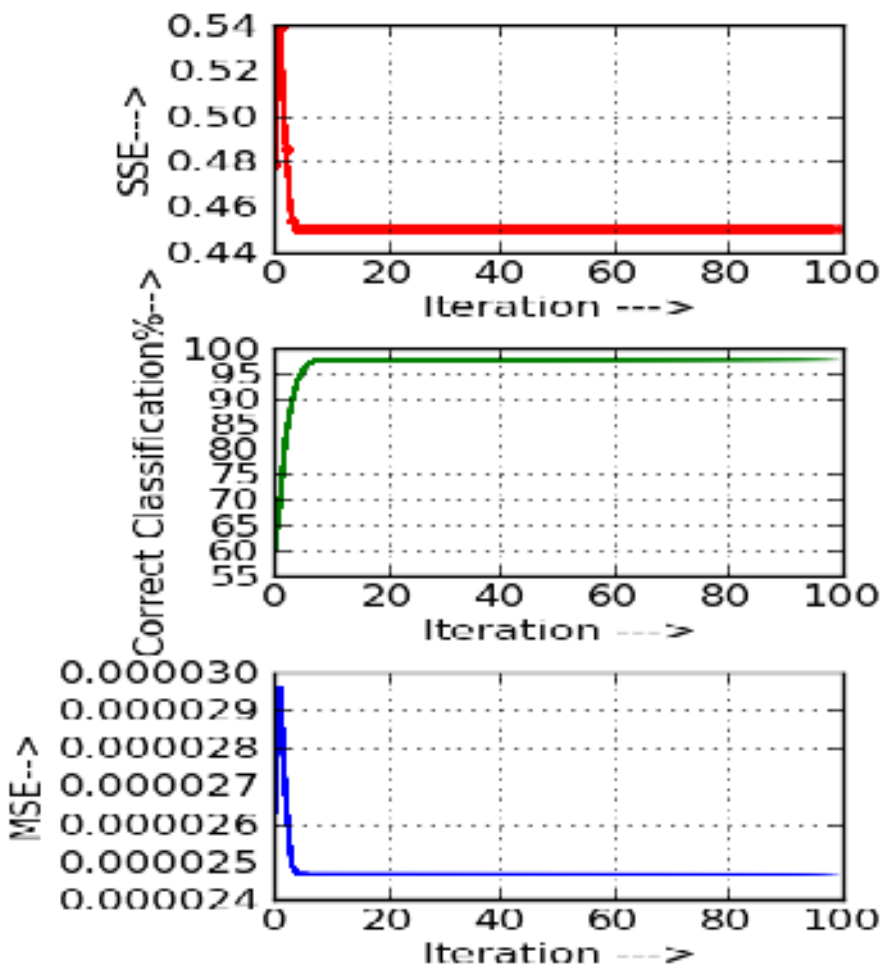

Figure 4. Iris data set training with ABCPNN

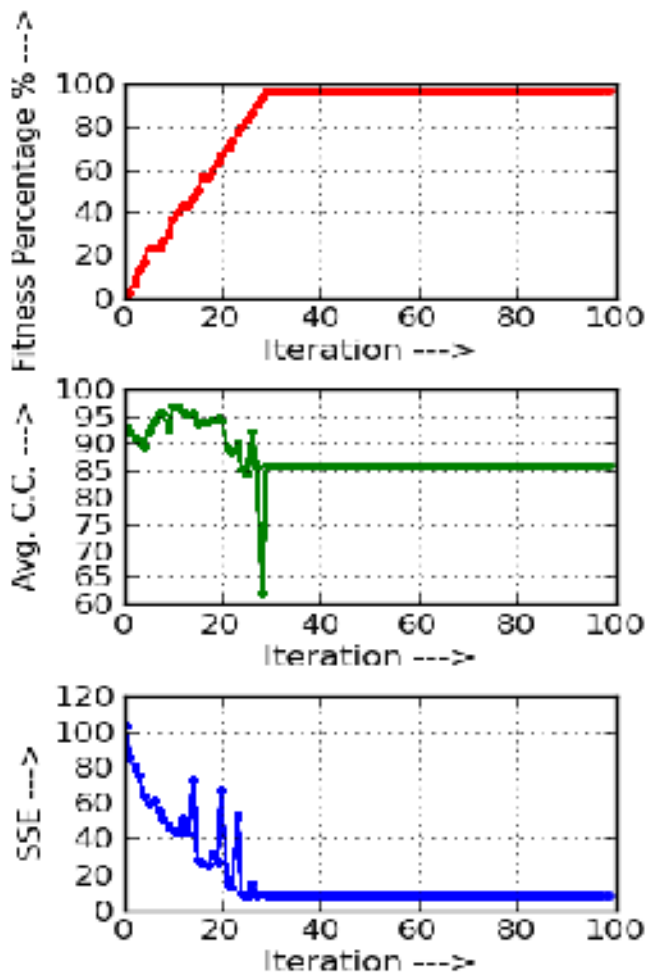

Figure 5. Iris data set training withGABPNN

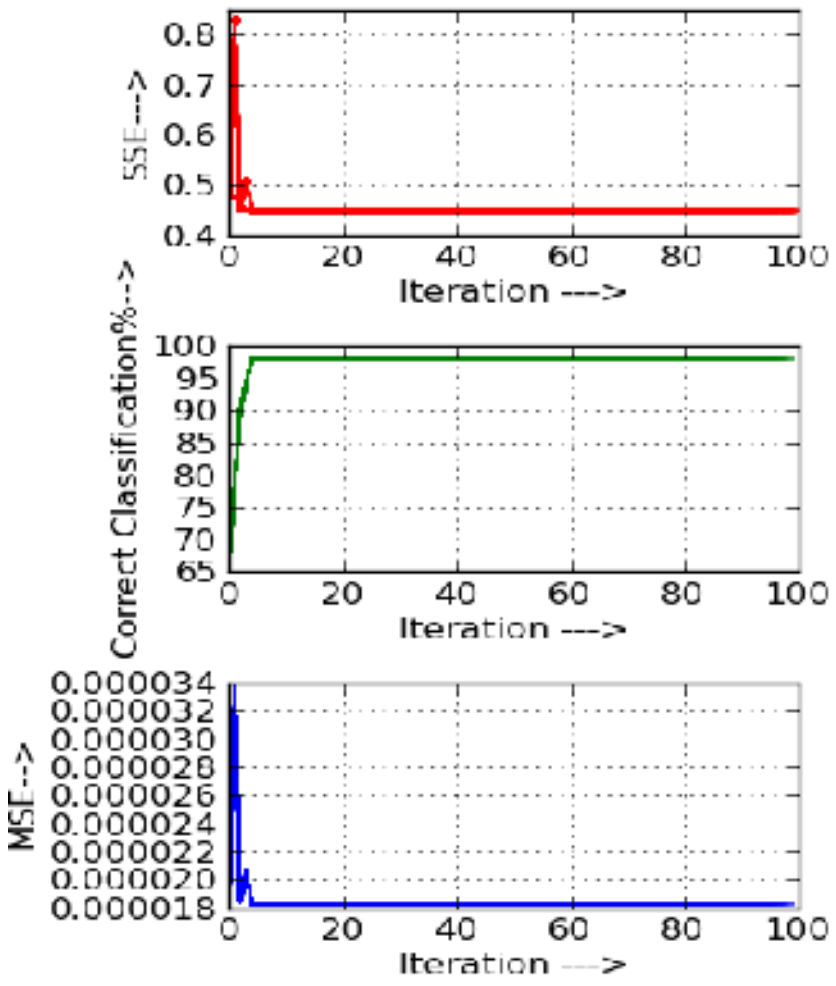

Figure 6. Wine  data set training with ABCBPNN

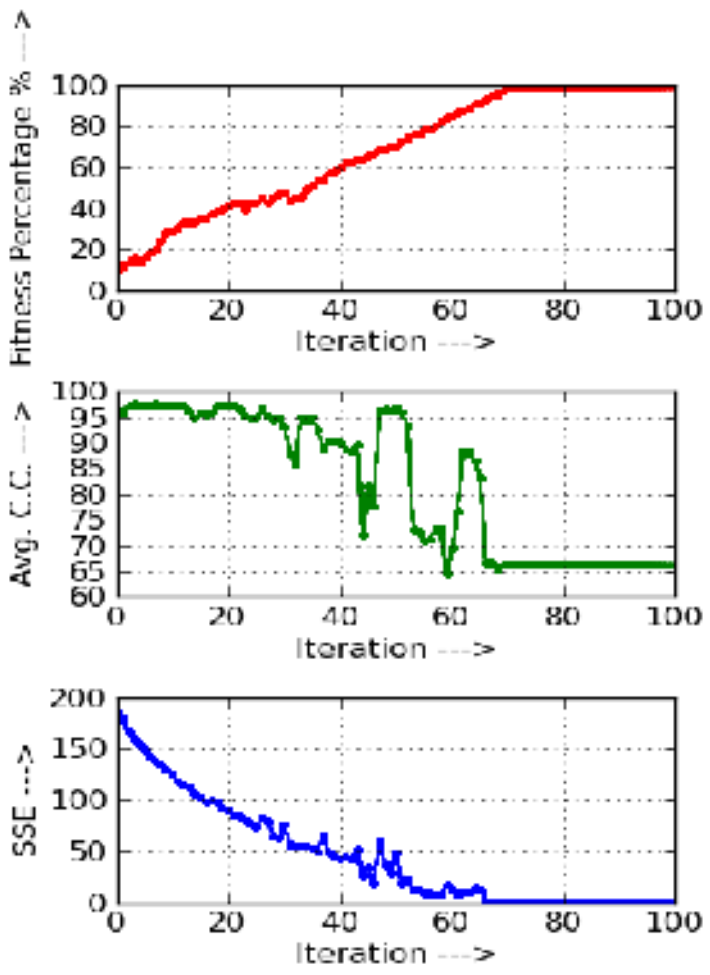

Figure 7. Wine data set training with GABPNN

The experiment with iris data set is depicted in Figure 4. and Figure 5. It is clearly observed from the figure that the speed of convergence is also improved for proposed method . In the wine data set training the improvement is observed in training with the proposed method. The statistics of training is in the table1. In the training with wine data, the convergence rate and the speed of the proposed method is better than that of the GABPNN. Figure 6 and Figure 7 is represented the training with wine data sets. The soybean data sets training is represented by the Figure 8 and Figure 9. The comparative pictorial presentation shows the improvement in the training with proposed method. The soyaben stable correct classification rate for GABPNN is absent in the Table 1 because it is observed that correct classification rate of genetic algorithm is not stable within 100 iteration. The proposed method is produced 91.5% correct classification rate. The glass data set is also analysed with the proposed method and genetic algorithm based back-propagation method. In Figure 10 and Figure 11 the training data of glass data set are plotted.

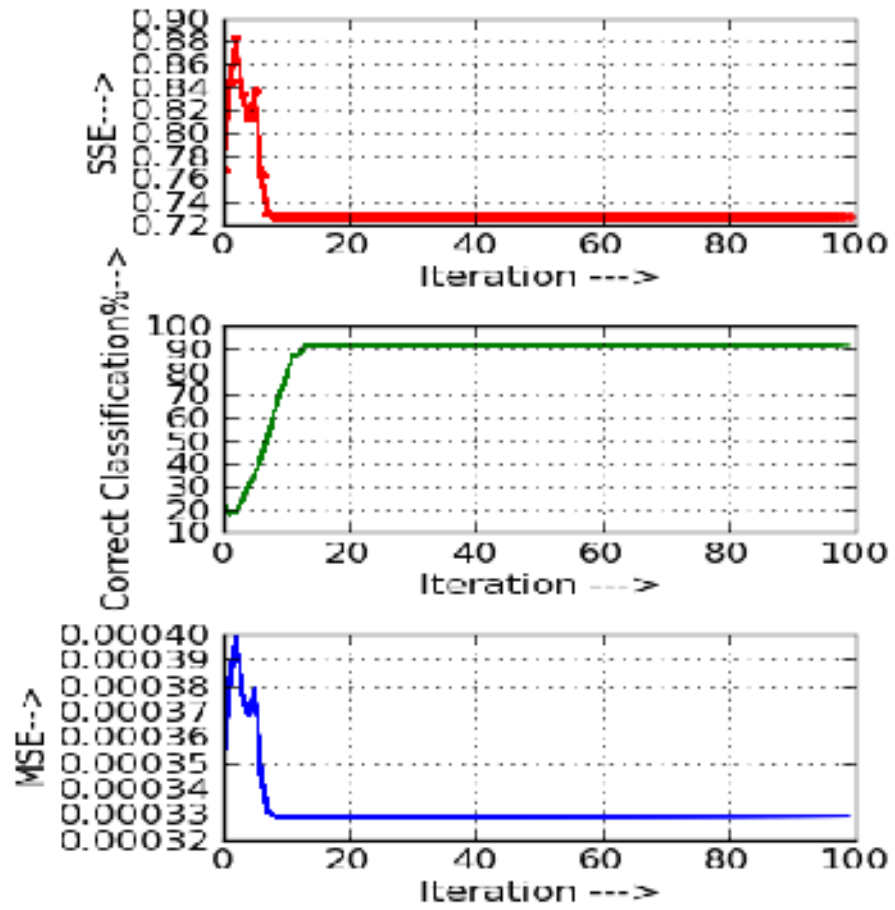

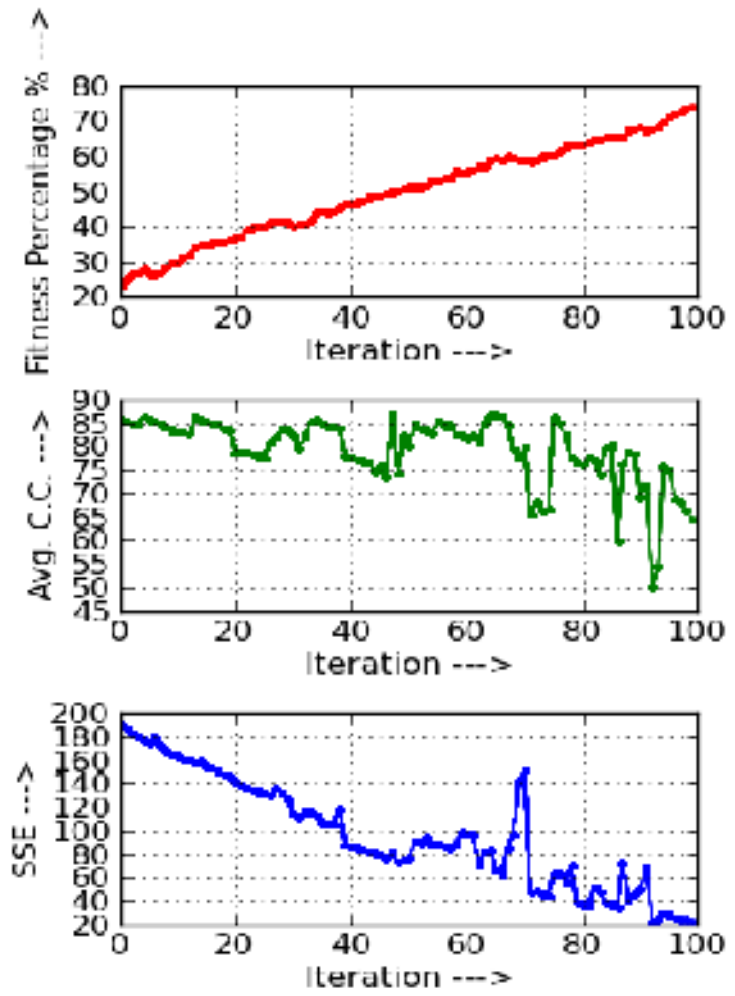

Figure 8: Soybean data set training with ABCBPNN

Figure 9: Soyabean data set training with GABPNN

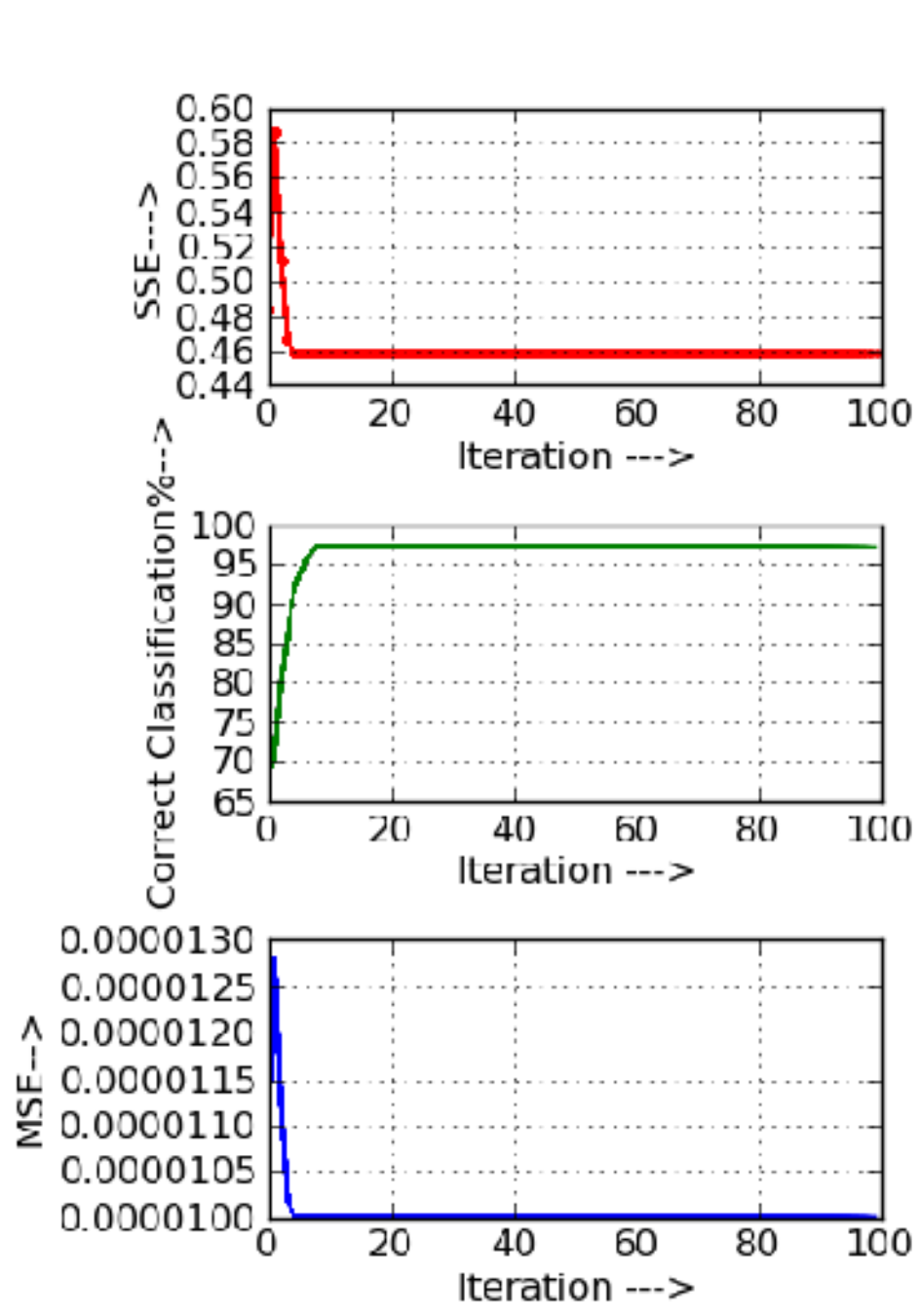

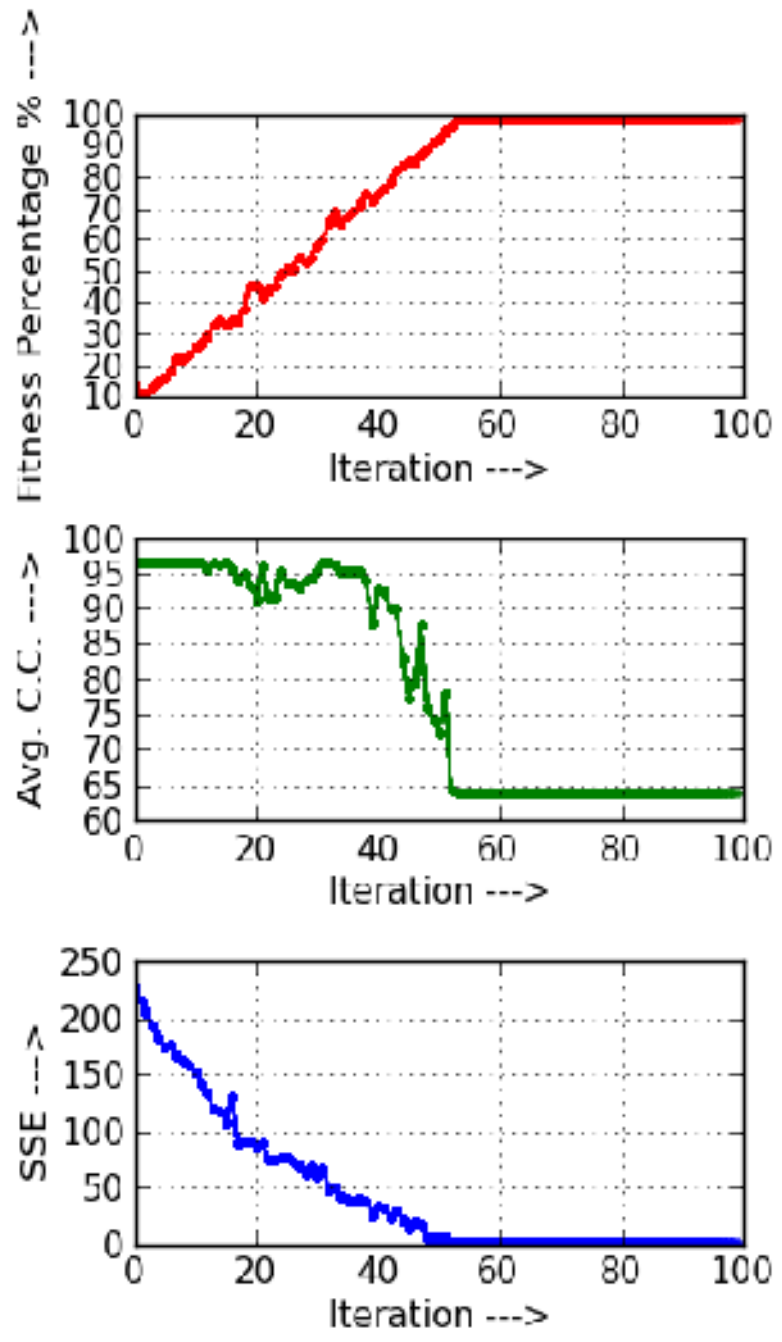

Figure 10. Glass data set training with ABCBPNN

Figure 11. Glass data set training with GABPNN

Table 1 shows the correct classification rate for GABPNN algorithm which is stable in the 64% but the maximum classification rate is 96.7. The proposed method is produced an improved classification rate for the glass data set and the stable classification rate is 97.2%.

Table1. Analysis of performance on four data sets.

| Data set | Algorithm Name | SSE | Correct Classification | | |
|---|---|---|---|---|---|
| | | | MAX | MIN | Stable |
| Iris | GABPNN | 8.29 | 96.96 | 61.93 | 85.93 |
| | **ABCBPNN** | **0.44** | **97.78** | **59.75** | **97.78** |
| Wine | GABPNN | 0.51 | 97.4 | 64.58 | 66.24 |
| | **ABCBPNN** | **0.44** | **98.1** | **67.73** | **98.1** |
| SoyBean | GABPNN | 28.08 | 86.8 | 49.87 | Not Stable in 100 iterations |
| | **ABCBPNN** | **0.72** | **91.5** | **17.87** | **91.5** |
| Glass | GABPNN | 0.51 | 96.7 | 64 | 64 |
| | **ABCBPNN** | **0.45** | **97.2** | **67.96** | **97.2** |

## 4. CONCLUSIONS

Artificial bee colony based back-propagation algorithm is the newly proposed algorithm to optimize the back-propagation neural network training. Conventional artificial bee colony algorithm is modified in terms of movement of employed bee to the food source and the movement of employed bee is used to modify the weight and bias of each solution towards optimum solution. The involvement of scout bee in each iteration of back-propagation optimization phase are decided according to the average performance of solutions and thus the process of exploitation is maintained in the proposed method. The performance of artificial bee colony based back-propagation (ABCBPNN) is compared with the genetic algorithm based back-propagation neural network training algorithm. The performance is analysed on the basis of 1. sum of squared error or solution quality 2. convergence speed 3. stability on the optimum solution. It is observed that the algorithm outperform the compared one over all of the tested problems and hence the efficiency of the proposed method gets established to be improved.

## REFERENCES

[1] J.H. Holland,( 1975) “Adaptation in Natural and Artificial Systems”, University of Michigan Press, Ann Arbor, MI.

[2] Dorigo M. & Stützle T.,(2004) “Aant Colony optimization”, MIT Press, Cambridge.

[3] K.V. Price, R.M. Storn, J.A. Lampinen (Eds.),(2005) “Differential Evolution: A Practical Approach to Global Optimization”, Springer Natural Computing Series.

[4] J. Kennedy, R.C. Eberhart, (1995) “Particle swarm optimization”, Proceedings of the IEEE International Conference on Neural Networks, vol. 4, pp. 1942–1948.

[5] X.S. Yang, (2005) “Engineering Optimizations via Nature-Inspired Virtual Bee Algorithms”, Lecture Notes in Computer Science, 3562, Springer-Verlag GmbH, pp. 317.

[6] David J. Montana and Lawrence Davis, (1989) "Training a Feed-forward Neural Networks using Genetic Algorithms", Journal of Machine Learning, pp. 762-767.

[7] N. Chakraborti, A. Kumar, (2003)“The optimal scheduling of a reversing strip mill: studies using multi-population genetic algorithms and differential evolution”, Mater. Manuf. Processes 18, pp.433–445.

[8] N. Chakraborti, P. Mishra, S. ErkocÂ¸, (2004) “A study of the Cu clusters using gray-coded genetic algorithms and differential evolution”, J. Phase Equilib. Diffus. 25, pp.16–21.

[9] Y. Fukuyama, S. Takayama, Y. Nakanishi, H. Yoshida, (1999)"A particle swarm optimization for reactive power and voltage control in electric power systems", Genetic and Evolutionary Computation Conference,pp. 1523–1528.

[10] Guang-Feng Deng & Woo-Tsong lin, (2011) " Ant colony optimization-based algorithm for airline crew scheduling problem", International Journal of Expert System with Applications, 38, pp. 5787-5793.

[11] S. Nakrani, and C. Tovey,(2004) "On Honey Bees and Dynamic Allocation in an Internet Server Colony," Proceedings of 2nd International Workshop on the Mathematics and Algorithms of Social Insects, Atlanta, Georgia, USA.

[12] G.Z. Markovic, D. Teodorovic, and V.S. Acimovic-Raspopovic, (2007) "Routing and Wavelength Assignment in All-Optical Networks Based on the Bee Colony Optimization," AI Communications – The European Journal on Artificial Intelligence, volume 20(4) pp. 273-285.

[13] H.F. Wedde, M. Farooq, T. Pannenbaecker, B. Vogel, C. Mueller, J. Meth, and K. Jeruschkat,(2005) "BeeAdHoc: An energy efficient routing algorithm for mobile ad hoc networks inspired by bee behavior," GECCO 2005, Washington DC, USA.

[14] Samad, T., 1990, "Back-propagation improvements based on heuristic arguments", Proceedings of International Joint Conference on Neural Networks, Washington, 1, pp. 565-568

[15] Sperduti, A. & Starita, A. , 1993, "Speed up learning and network optimization with Extended Back-propagation", Neural Networks, 6, pp. 365-383.

[16] Van Ooten A. , Nienhuis B, 1992, "Improving the convergence of the back-propagation algorithm", Neural Networks, 5, pp. 465-471.

[17] C. Charalambous, 1992, "Conjugate gradient algorithm for efficent training of neural networks", IEEE Procedings-G , vol. 139, 3.

[18] Levenberg, K., 1944, "A method for the solution of certain problem in least squares", Quart. Appl. Math., 2, pp.164-168.

[19] Marquardt , D. , 1963, " An algorithm for least sqare estimation of nonlinear parameters", SIAM J. Appl. Math., 11, pp. 431-441.

[20] M. T. Hagan and M. B. Menhaj, 1994, "Training feedforward networks with the Marquardt algorithm," IEEE Trans. Neural Netw., vol. 5, no. 6, pp.989–993.

[21] Celal Ozturk & Devis Karaboga, (2011) " Hybrid artificial bee colony algorithm for neural network training", IEEE Congress on Evolutionary Computing, pp. 84-88.

[22] P. Y. Kumbhar & S. Krishnan, (2011), "Use of artificial bee colony algorithm in artificial neural network synthesis", International Journal of Advanced Engineering Sciences and Technologies, Vol. No. 11, 1, pp. 162-171.

[23] J. Zhang, J. Zhang, T. Lok & M.Lyu, (2007) " A hybrid particle swarm optimization backpropagation algorithm for feed forward neural network training" , Applied Mathematics and Computation, Vol. 185, pp. 1026-1037.

[24] Michael S. Engel (1999). "The taxonomy of recent and fossil honey bees (Hymenoptera: Apidae: Apis)". Journal of Hymenoptera Research 8: 165–196

[25] UCI Machine Learning Repository : Iris Data Set - http://archive.ics.uci.edu/ml/datasets/Iris

[26] UCI Machine Learning Repository : Wine Data Set - http://archive.ics.uci.edu/ml/datasets/Wine

[27] UCI Machine Learning Repository : Glass Identification Data Set http://archive.ics.uci.edu/ml/datasets/Glass+Identification

[28] UCI Machine Learning Repository : Soybean Data Set - http://archive.ics.uci.edu/ml/datasets/Soybean+%28Small%29

## Authors

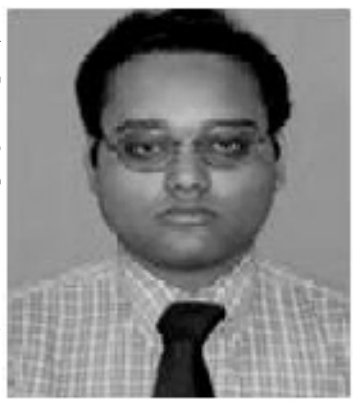

**Sudarshan Nandy** was born on 6thday of March 1983. He completed B.Tech in Computer Science and Engineering from Utkal University,and M.Tech in Computer Science and Engineering from West Bengal University of Technology in the years 2004 and 2007 respectively. He has been serving as Assistant Professor in JIS College of Engineering, since 2009. He is pursuing his Ph.D work under the guidance of Prof. Partha Pratim Sarkar and Prof. Achintya Das. His area of interest is Computational Intelligence, Web Intelligence, Meta-heuristics algorithm and Neural Network
.

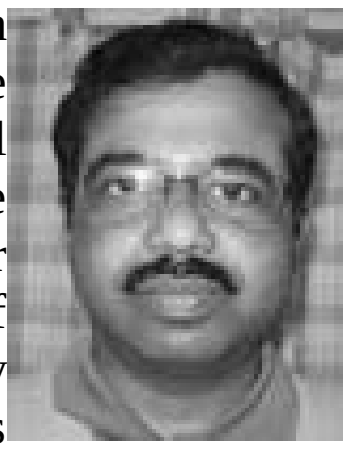

**Dr. Partha Pratim Sarkar** obtained his Ph.D in engineering from Jadavpur University in the year 2002. He has obtained his M.E from Jadavpur University in the year 1994. He earned his B.E degree in Electronics and Telecommunication Engineering from Bengal Engineering College (Presently known as Bengal Engineering andScience University,Shibpur) in the year 1991. He is presently working as Senior Scientific Officer (Professor Rank) at the Dept. of Engineering & Technolog ical Studies, University of Kalyani. His area of research includes, Microstrip Antenna, Microstrip Filter, Frequency Selective Surfaces, and Artificial Neural Network. He has contributed to numerous research articles in various journals and conferences of repute. He is also a life Fellow of IETE.

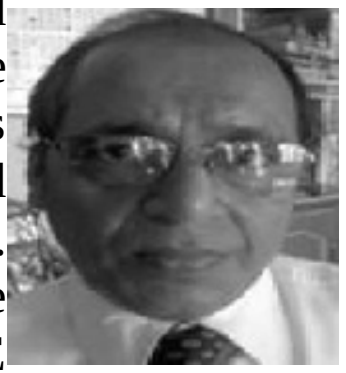

Dr. Achintya Das was born on 8th February 1957. He completed B.Tech., M.Tech and Ph.D (Tech) in the subject of Radio Physics Electronics from Calcutta University in the years of 1978, 1982 and 1996 respectively. He served as Executive Engineer in Philips from 1982 to 1996. He is Professor and Head of the department of Electronics and Communication Engineering of Kalyani Govt. Engineering College, Kalyani, West Bengal. He also worked for twelve years as Visiting Professor at Calcutta University. He has more than forty research publications so far. He is fellow members (life) of IE and IETE professional bodies.